\documentclass[sn-basic]{sn-jnl}

\usepackage{graphicx}
\usepackage{multirow}
\usepackage{amsmath,amssymb,amsfonts}
\usepackage{amsthm}
\usepackage{mathrsfs}
\usepackage[title]{appendix}
\usepackage{xcolor}
\usepackage{textcomp}
\usepackage{manyfoot}
\usepackage{booktabs}
\usepackage{algorithm}
\usepackage{algorithmicx}
\usepackage{algpseudocode}
\usepackage{listings}
\usepackage{colortbl}
\usepackage{rotating}

\usepackage[normalem]{ulem}
\useunder{\uline}{\ul}{}

\begin{document}

\title[Article Title]{A Comparison of Deep Learning and Established Methods for Calf Behaviour Monitoring}


\author*[1,2]{\fnm{Oshana} \sur{Dissanayake}}\email{oshana.dissanayake@ucdconnect.ie}

\author[1,2,5]{\fnm{Lucile} \sur{Riaboff}}\email{lucile.riaboff@inrae.fr}

\author[2,3,4]{\fnm{Sarah E.} \sur{McPherson}}\email{sarah.McPherson@teagasc.ie}

\author[2,3]{\fnm{Emer} \sur{Kennedy}}\email{emer.Kennedy@teagasc.ie}

\author[1]{\fnm{P\'{a}draig} \sur{Cunningham}}\email{padraig.cunningham@ucd.ie}

\affil[1]{\orgdiv{School of Computer Science}, \orgname{University College Dublin}, \orgaddress{\street{Belfield}, \city{Dublin}, \postcode{D04 V1W8}, 
\country{Ireland}}}

\affil[2]{\orgdiv{}, \orgname{VistaMilk SFI Research Centre}, \orgaddress{
\country{Ireland}}}

\affil[3]{\orgdiv{Animal \& Grassland Research and Innovation Centre}, \orgname{Teagasc}, \orgaddress{\street{Moorepark, Fermoy}, \city{Co. Cork}, \postcode{P61C997}, 
\country{Ireland}}}

\affil[4]{\orgdiv{Animal Production Systems Group}, \orgname{Wageningen University}, 
\orgaddress{
\city{Wageningen}, 
\country{The Netherlands}}}

\affil[5]{\orgdiv{GenPhySE}, \orgname{Université de Toulouse}, \city{INRAE, ENVT}, \postcode{31326}, \state{Castanet-Tolosan}, \country{France}}


\abstract{In recent years there has been considerable progress in research on human activity recognition using data from wearable sensors. This technology also has potential in an animal welfare context in livestock science. In this paper we report on research on animal activity recognition in support of welfare monitoring. The data comes from collar-mounted accelerometer sensors worn by Holstein and Jersey calves; the objective being to detect changes in behaviour indicating sickness or stress. A key requirement in detecting changes in behaviour is to be able to classify activities into classes such as drinking, running or walking. In Machine Learning terms, this is a time-series classification task and in recent years the Rocket family of methods have emerged as the state-of-the-art in this area. For our analysis we have over 27 hours of labelled time-series data from 30 calves. We present the performance of Rocket on a 6-class classification task on this data as a baseline. Then we compare this against the performance of 11 Deep Learning (DL) methods that have been proposed as promising methods for time-series classification. Given the success of DL  in related areas it is reasonable to expect that these methods would perform well here as well. Surprisingly, despite taking care to ensure that the DL methods are configured correctly, none of them match the performance of Rocket. A possible explanation for the impressive success of Rocket is that it has the data encoding benefits of DL models in a much simpler classification framework. }




\maketitle
 
\section{Introduction}\label{sec_introduction}
Machine Learning (ML) and Deep Learning (DL) techniques have the potential to transform many application areas by excelling in complex pattern detection. In livestock sciences, these techniques may provide models that accurately classify and predict animal behaviours, enhancing our understanding of animal welfare and optimizing livestock management practices. Specifically, ML and DL algorithms are effective for processing and analyzing large datasets to identify patterns and anomalies, making them ideal for animal management applications. In this paper we address the challenge of using ML/DL models to recognize specific activities such as drinking milk, grooming, lying, walking, and running in young calves. This can provide  insights into the health, stress levels, and overall well-being.

Behaviour is a critical indicator of the stressors a calf is experiencing \citep{neamț2019weaning, van2005responses}. Thus, tracking calf behaviour can significantly aid in developing an automated stressor identification system. Sensors have long been used for behaviour monitoring, yielding remarkable results in the Human Activity Recognition (HAR) field \citep{bayat2014study, ignatov2018real, lee2017human}. In Animal Activity Recognition (AAR), accelerometer sensors are widely used as well \citep{kleanthous2022deep, tran2021iot, chakravarty2019novel}. The dataset used in this study, ActBeCalf, is a product of behaviour monitoring using accelerometer sensors \citep{dissanayake2024accelerometer}. Modern ML/DL approaches efficiently handle large amounts of data, making sensor data suitable for generating better accuracies.

While there has already been considerable progress in HAR, this field is still developing for animals and is at a very early stage for calf behaviour classification. Most related studies classify a limited number of behaviours, and the robustness of the developed models is often questionable \citep{riaboff2022predicting}.

Calves are prone to more diseases and stressors being at an early stage in their lives. However, the most common commercial practice is to
wean the calf at birth, with restrictions on milk and housing in separate pens \citep{whalin2021understanding}. Besides, calves undergo stressors during transportation \citep{mackenzie1997effect},  dehorning \citep{faulkner2000reducing}, social isolation \citep{whalin2021understanding}, etc. Thus, a system capable of identifying these stressors and preventing health issues or deaths due to their effects is essential \citep{koknaroglu2013animal}. Such a system will improve animal welfare and connect farming with societal expectations as consumers increasingly value the practices applied in livestock production. \citep{cardoso2016imagining}. 

To address the above issues, we have developed a methodology which includes a data separation mechanism ensuring the generalisation of the models and considering a wide range of behaviour classes, including a class known as `other', which composes 19 different behaviours to ensure the robustness of the model for unseen data. The study compares the performance of 11 DL methodologies in the Time Series Classification (TSC) field against ROCKET's performance with RidgeClassifier, which has proved to be the most effective in maintaining these criteria \citep{dissanayake2024evaluating}. 
Because the performance of the DL methods is surprisingly disappointing we also present an evaluation on the WISDM dataset (see section \ref{subsec_model_validation}), which is widely used in the time series classification field. This validation of the DL algorithms indicates that they perform as expected, confirming that our configurations were correct.

The contributions of this paper can be summarised as follows:
\begin{itemize}
    \item \textbf{Comparative Analysis of Deep Learning Methodologies: }We evaluate 11 DL methods for time series classification against the ROCKET (with a Ridge Classifier). The study finds ROCKET to be the most effective in maintaining generalisation and robustness for unseen data.
    \item \textbf{Use of ActBeCalf Dataset and Generalisation Strategy: } We use the ActBeCalf dataset for behaviour monitoring and employ a data separation mechanism to ensure the generalisation of models. We find that including a composite `other' class comprising 19 different behaviours enhances the model's robustness.
\end{itemize}

This paper is organised as follows: Section 2 discusses calf welfare and related work. Section 3 discusses the use of machine learning in calf behaviour classification and describes in depth the ROCKET and DL algorithms used in this study. Section 4 describes the ActBeCalf dataset and methodology for classification. Section 5 presents the results and discuss the performance of the algorithms. From Section 7, the paper is concluded.

\section{Calf Welfare}\label{sec_calf_welfare}
Behaviour is one of the critical indicators of stress in calves. Behaviour changes, such as decreased activity bouts, altered feeding patterns, social isolation and increased lying down time, can indicate a stressful episode \citep{mahendran2023calf, nikkhah2023understanding}. Monitoring these behavioural changes can provide early warning signs of physiological and psychological health issues \citep{dissanayake2022identification}, helping to make early interventions to prevent long-term effects on the calf's health. It will help reduce costs, time, and labour associated with treating a much more intense stressor. It can also significantly aid management in recommending and creating routine practices for less stressful animals and promote calf welfare \citep{mcpherson2022effect}. However, such a system would require continuous behaviour monitoring to detect the changes, which is highly time-consuming, labour-intensive and deemed impractical for humans \citep{penning1983technique}. 

Thus, an automated behaviour monitoring system is required. Sensors are becoming widely popular in the field of behaviour monitoring in both animals and humans. In that regard, sensors such as accelerometers, gyroscopes and GPS trackers are recommended \citep{jiang2023precision, rushen2012automated}. Accelerometer sensors can monitor motion in 3 dimensions and provide a continuous data stream based on frequency, capturing minute motions within significantly smaller time windows. They also can function in highly varying environmental conditions with long battery operating times and can easily be integrated into different monitoring systems such as neck collars or leg-mounted devices. This explains their rapid adoption in livestock farming \citep{chakravarty2019novel, martiskainen2009cow, moreau2009use, tatler2018high, yu2023accelerometer}. 

Calf behaviour classification using accelerometer sensors and machine learning algorithms has been successful for some easy-to-detect behaviours.
However, there are a few challenges with the classifications of less prominent behaviours, such as transitional and maintenance behaviours, and when the objective is to monitor multiple behaviours at the same time \citep{riaboff2022predicting}. Transitional behaviours (such as lying-down and rising) are observed less frequently and a sensitivity of less than 70\% has been reported \citep{martiskainen2009cow,  vazquez2015classification}. Maintenance behaviours such as urinating, drinking, social interaction, and self-grooming tend to score accuracies less than 80\% \citep{lush2018classification, rodriguez2020identifying}. \citet{dissanayake2024evaluating} and \citet{hosseininoorbin2021deep} have shown that calf behaviour classification for a smaller number of classes yields excellent performance, but the performance decreases substantially when the classification considers more than five classes. However, less prominent behaviours (transitional behaviours, social interaction) compared with behaviours like lying down, standing, eating, drinking, etc., also have a say in the calf's health. Thus, it is important to identify such behaviours as well. Several clear behavioural changes can be observed after stressful processes like weaning and dehorning. An increase in behaviours like seeking and walking and a reduction in behaviours like playing can be observed after weaning \citep{enriquez2010effects}. More lying-down and standing-up transitions, more scratching and less self-grooming behaviours are observed after dehorning \citep{morisse1995effect}.

\section{Machine Learning Methods for Time Series Classification}\label{sec_ml_tech_for_ts}
A significant gap in much of the related work is the lack of a generic approach in the machine learning models used for classification \citep{Brownlee2020DataPreparation, kuhn2014futility}. Usually, the datasets are created as a whole from the data from all the subjects, and a random split is made for training and testing data. There is no separation of the data between the subjects; the data from the same animal can be seen in both the train and test data. This drastically reduces the generic nature of the model, although the reported accuracy results are high for that particular dataset. Thus, when the model is tested with new animal data, the performance is low due to the dissimilarity between the feature spaces of the training and test sets \citep{rahman2018cattle}. 

Both DL and ML methods have been used in the studies on calf behaviour classification. \citet{hernandez2024machine} used a GNSS sensor, 9-axis IMU and a Digital thermometer to classify cattle behaviour into general behaviours (4 behaviours): grazing-eating, ruminating, neutral, walking and (2 behaviours): standing and lying. Classifiers, including SVM, decision trees, and RF, have been tested, and the extra-randomized trees and SVMs have been the most efficient. As per the results, 57\% accuracy was achieved for general behaviour (4 classes) and 85\% for standing behaviour (2 classes). \citet{carslake2020machine} have used an AdaBoost ensemble learning algorithm to classify two postures (standing and lying) and seven behaviours including ruminating, self-grooming, and locomotor play in Holstein dairy calves. They report an overall accuracy of 94.38\% using a 4-second window for posture classification, and behaviour classification has yielded an overall accuracy of 95.72\% using a 3-second window. \citet{cantor2022using} have used a $k$-Nearest Neighbor ($k$NN) algorithm to study the possibility of early detection of clinical Bovine Respiratory Disease (BRD) status in calves based on behavioural patterns and health data. For preclinical BRD scenarios, the algorithm was highly accurate (over 90\%) up to 6 days prior to clinical BRD diagnosis using automated features. \citet{vazquez2024quantification} have also used an AdaBoost ensemble learning algorithm to classify calves' play and non-play behaviours and achieved an overall accuracy greater than 94\%. \citet{zhang2024analysis} have classified between healthy calves and sick calves (with diarrhoea) using YOLOv8n deep learning model. Features such as standing time, lying time, number of lying bouts, and average bout duration (using video feeds) have been used for the model training, and the model achieved a mean average precision of 0.995. 

A comparative study has been conducted by \citet{el2024comparative} 
between Random Forest (RF), Support Vector Machine (SVM) and Convolutional Neural Network (CNN) to classify the cattle behaviour into eating, ruminating and other classes. CNN achieved the highest accuracy of 95\%, surpassing RF and SVM (72\% and 83\%, respectively). \citet{arablouei2023animal} have used the grazing beef cattle (Angus breed) to classify their behaviour into grazing, walking, ruminating, resting, drinking, and other using a Deep Neural Network (DNN)-based algorithm. The algorithm demonstrated excellent intra- and inter-dataset classification accuracy. Specific results include high Matthews correlation coefficient (MCC) values across different datasets, indicating effective behaviour classification.

\subsection{ROCKET}\label{subsec_rocket}

\textbf{R}and\textbf{O}m \textbf{C}onvolutional \textbf{KE}rnel \textbf{T}ransform is a feature extraction mechanism inspired by the deep learning feature extraction mechanism, first introduced by \citep{dempster2020rocket}. ROCKET uses a large number of random convolutional kernels (default = 10,000) to extract features from time series data. These kernels have randomly generated parameters allowing the algorithm to capture a wide variety of patterns and characteristics inherent in the data. These are:

\begin{itemize}
    \item \textbf{lengths}: Randomly chosen from 7, 9, or 11 with equal probability.
    \item \textbf{weights}: Sampled from a normal distribution and then mean-centered.
    \item \textbf{biases}: Sampled from a uniform distribution.
    \item \textbf{dilations}: Sampled on an exponential scale.
    \item \textbf{paddings}: Applied randomly with equal probability to handle edge effects.
\end{itemize}
By applying these kernels to the time-series data, ROCKET creates a high-dimensional feature vector (the amount of features depends on the number of kernels) to be used by the classification algorithms. It extracts two features per kernel: 
\begin{enumerate}
    \item \textbf{Maximum Value}: The highest value from the convolution operation.
    \item \textbf{Proportion of Positive Values (PPV)}: The proportion of the feature map that contains positive values. 
\end{enumerate}

ROCKET is famous for its efficiency and its accuracy. Using random kernels means there is no need to train a complex network, significantly reducing the computational overhead. Due to its efficient implementation, ROCKET scales well with increasing data sizes and dimensions. It can easily handle large and complex time series datasets, making it suitable for real-world applications where data volumes can be substantial. The simplicity of the feature space extracted by ROCKET enables the application of straightforward linear classifiers, such as ridge regression. This integration merges the interpretability and simplicity of linear models with ROCKET's robust feature extraction capabilities. One of the key advantages of ROCKET is that it requires minimal hyperparameter tuning. The randomness in the kernel generation process inherently provides a diverse set of features, reducing the need for extensive parameter optimization.

\textbf{MiniRocket} (Mini RandOm Convolutional KErnel Transform), also introduced by \citep{dempster2021minirocket}, is a streamlined variant of the ROCKET algorithm with increased speed and efficiency for time series classification. MiniRocket uses a simpler, fixed set of kernels (84 kernels of length 9 with weights restricted to two values -1 and 2), making the process almost entirely deterministic. This change significantly reduces the randomness involved and increases computational speed. Moreover, only one feature, \textit{PPV}, is extracted per kernel. This methodological shift enables MiniRocket to achieve similar accuracy levels to ROCKET while being up to 75 times faster, especially on larger datasets. The work presented here uses MiniRocket. 

\textbf{Ridge Classifier} is a linear classifier that works well with ROCKET because, as a linear model, it works well in a high-dimension feature space. We have used RidgeClassifierCV, the scikit-learn implementation, which incorporates cross-validation for regularization. \citet{dempster2020rocket} recommend using a ridge regression classifier in situations where there are fewer training examples than features. Its built-in regularization helps prevent overfitting, effectively managing the high-dimensional feature space generated by the method. Also, the automatic cross-validation tuning of the regularization parameter ensures optimal model performance without extensive manual hyperparameter tuning. Additionally, RidgeClassifierCV is computationally efficient and scales well with large datasets, making it ideal for handling the substantial number of features produced by ROCKET.

\subsection{Deep Learning Methods}\label{subsec_dl_methods}

We have considered 10 deep learning algorithms proposed by \cite{ismail2019deep} for this study. Despite their success in other domains like computer vision and natural language processing, the authors highlight the surprising underutilization of deep neural networks (DNNs) in TSC. They present a comprehensive empirical study comparing various state-of-the-art DNN architectures, including Convolutional Neural Networks (CNNs) and Residual Networks (ResNets), by training over 8,730 models on 97 UCR/UEA archive datasets and 12 multivariate datasets. The methods are adjusted to use the generalisation mechanism used by out study. The DL implementations used, and our evaluation code is available in a GitHub repository \footnote{https://github.com/Oshana/ESWA}. 

We have also considered the ConvTran algorithm introduced by \cite{ConvTran2023} because of the impressive performance on benchmarks datasets. The ConvTran model is an multivariate time series classification framework that integrates convolutional neural networks (CNNs) and transformers. It introduces time Absolute Position Encoding (tAPE) and efficient Relative Position Encoding (eRPE) to enhance position and data embedding. The model starts with convolutional layers to capture local temporal patterns, followed by transformer blocks to handle long-range dependencies. 

Before presenting an overview of the 10 DL models and ConvTran, here are explanations of some terms used:

\begin{itemize}
    \item \textbf{Sigmoid Activation: }The sigmoid activation function maps input values to an output range between 0 and 1, as shown in Equation \ref{eq:sigmoid}. This function is particularly useful for binary classification tasks as it outputs a probability-like value, facilitating the interpretation of results \citep{Rumelhart1986}. However, the sigmoid function can suffer from the vanishing gradient problem, where gradients become too small for effective learning in DNNs.

    \begin{equation} \label{eq:sigmoid}
        \sigma = \frac{1}{1+e^{-x}}
    \end{equation}

    \vspace{0.7em}

    \item \textbf{Softmax Activation: }The softmax activation function converts raw prediction scores into probabilities that sum to one, making it ideal for multi-class classification problems. Softmax operates by exponentiating each output and then normalizing by the sum of all exponentiated outputs, effectively highlighting the most likely class while suppressing others \citep{bridle1989training}. This approach helps in interpreting the output of a model as a probability distribution over possible classes.

    \vspace{0.7em}
    
    \item \textbf{ReLU Activation: }
    An activation function introduces non-linearity to a DL model which is important to help the model learn complex patterns and functions in the data. ReLu is an efficient activation function due to its simplicity, which adds non-linearity to the model without raising computational complexity too much \citep{nair2010rectified}. Mathematically, ReLU is defined as as shown in Equation \ref{eq:ReLu}. This means that the output is zero if the input is negative, and it is equal to the input if the input is positive.
    \begin{equation} \label{eq:ReLu}
        f(x) = max(0,x)
    \end{equation}

    \item \textbf{PReLU Activation: } The Parametric Rectified Linear Unit (PReLU) is an advanced activation function that introduces a learnable parameter to the traditional ReLU, allowing for adaptive control over the slope of the negative part of the input \citep{he2015delving}. This adaptability helps mitigate the problem of dying ReLUs, where neurons become inactive and stop learning, thus enhancing the overall performance of DNNs. 
    
    \vspace{0.7em}

    \item \textbf{GELU Activation: }The Gaussian Error Linear Unit (GELU) is an activation function that combines the properties of linear and non-linear activations by applying a smooth and differentiable Gaussian cumulative distribution function to the input \citep{hendrycks2016gaussian}. Unlike traditional activation functions like ReLU, GELU considers the input values' distribution, allowing for better handling of the input noise and improving model performance.
    
    \vspace{0.7em}
    
    \item \textbf{Adadelta Optimizer: }An optimizer is an algorithm used to adjust the weights of a neural network to minimize the loss function and improve the model's accuracy. Adadelta is an adaptive learning rate optimization algorithm that addresses the diminishing learning rates issue in AdaGrad by using exponential moving averages of squared gradients. This approach allows Adadelta to maintain more stable and effective learning rates throughout the training process, enhancing model convergence \citep{zeiler2012adadelta}.

    \vspace{0.7em}

    \item \textbf{Adam Optimizer: }The Adam optimizer, short for Adaptive Moment Estimation, combines the benefits of two other popular optimization algorithms: AdaGrad and RMSProp. It computes adaptive learning rates for each parameter by estimating the first and second moments of the gradients, which helps in faster convergence and better handling of sparse gradients \citep{kingma2014adam}. 

    \vspace{0.7em}

    \item \textbf{SGD Optimizer: }The Stochastic Gradient Descent (SGD) optimizer is a fundamental optimization algorithm in ML. It updates model parameters iteratively by taking steps proportional to the negative gradient of the loss function with respect to the parameters, calculated from randomly selected subsets of the training data \citep{Bottou2010}.

    \vspace{0.7em}
    
    \item \textbf{Dropout layers: }Dropout layers are a regularization technique used in neural networks to prevent overfitting by randomly setting a fraction of input units to zero during training. This forces the network to learn more robust features that are less reliant on specific neurons \citep{srivastava2014dropout}.

    \vspace{0.7em}
  
    \item \textbf{Hidden layers: }Hidden layers in neural networks are intermediate layers that process inputs from the input layer and pass the processed information to the output layer. They consist of neurons that apply activation functions to weighted inputs, enabling the network to learn complex patterns and representations \citep{nielsen2015neural}. The number and size of hidden layers can significantly influence the network's ability to model intricate data relationships.

    \vspace{0.7em}
    
    \item \textbf{Categorical cross-entropy: }Categorical cross-entropy is a loss function used in classification problems where the output is a probability distribution over multiple classes \citep{rusiecki2019trimmed}. It measures the difference between the true label distribution and the predicted probability distribution, penalizing incorrect classifications. This loss function is effective for training models helping to optimize the model's accuracy.
\end{itemize}

\subsubsection{Multi Layer Perceptron (MLP)}
A Multi-Layer Perceptron contains multiple layers of neurons. Each layer in an MLP is fully connected to the next, forming a network that can model complex patterns in data. The MLP consists of an input layer, one or more hidden layers, and an output layer. In this specific implementation, the MLP model processes data through three hidden layers with ReLU activation and dropout to prevent overfitting. The final layer uses softmax for class probabilities. It is compiled with categorical cross-entropy loss and optimized with Adadelta. Training involves backpropagation and learning rate reduction callbacks for efficiency.

\subsubsection{Fully Convolutional Neural Network (FCN)}
A Fully Connected Neural Network has the capability to handle sequential data such as time series. In this specific implementation, the FCN model consists of three Conv1D layers with 128, 256, and 128 filters, followed by batch normalization and ReLU activation. A global average pooling layer reduces spatial dimensions before the final dense layer, which uses softmax for class probabilities. The model is compiled with categorical cross-entropy loss and optimized with Adam. Training involves fine-tuning hyperparameters and using callbacks like learning rate reduction to enhance efficiency and performance based on validation loss.

\subsubsection{Residual Network (RESNET)}
A Residual Network can address the vanishing gradient problem to improve the training of very deep networks. In this specific implementation, the ResNet model has multiple residual blocks, each with three Conv1D layers (64, 128, and 256 filters), batch normalization, and ReLU activation. Residual connections enable easier training of deeper models. The final layer uses global average pooling and a dense layer with softmax for class probabilities. The model is compiled with categorical cross-entropy loss and optimized with Adam.

\subsubsection{Encoder}
The Encoder model leverages both Convolutional Neural Networks (CNN) and an attention mechanism to effectively capture spatial and temporal features in time series data. This implemented model has three convolutional blocks with Conv1D layers (128, 256, and 512 filters), instance normalization, PReLU activation, dropout, and max pooling. The output is split for attention processing, applying softmax to one part and multiplying it with the other to focus on relevant features. A dense layer with sigmoid activation follows, and the final layer uses softmax for class probabilities. The model is compiled with categorical cross-entropy loss and optimized with Adam.

\subsubsection{Multi-scale Convolutional Neural Network (MCNN)}
The Multi-Scale Convolutional Neural Network model is designed to handle time series data by capturing features at multiple scales. The implemented MCNN model preprocesses input data with downsampling and moving averages. It has multiple input branches, each with a Conv1D layer (256 filters, sigmoid activation) and max pooling. These branches are concatenated, followed by another Conv1D layer and max pooling. The output is flattened, passed through a dense layer, and the final layer uses softmax for class probabilities. The model is compiled with categorical cross-entropy loss and optimized with Adam. Training includes data augmentation and hyperparameter tuning.

\subsubsection{Time Le-Net (t-LeNet)}
The Time Le-Net model is designed to classify time series data using a combination of Convolutional Neural Networks (CNN) and data augmentation techniques. This implementation uses window warping and slicing for data augmentation. The CNN has two convolutional layers (5 and 20 filters) with ReLU activation and max pooling. The output is flattened and passed to a fully connected layer with 500 neurons, followed by a softmax classification layer. The model is compiled with categorical cross-entropy loss and optimized with Adam. Data augmentation increases training diversity, improving the model's ability to learn and classify complex time series patterns.

\subsubsection{Multi Channel Deep Convolutional Neural Network (MCDCNN)}
The Multi-Channel Deep Convolutional Neural Network is designed to classify multivariate time series data by leveraging the power of deep convolutional layers. This model uses separate convolutional pipelines for each variable in the multivariate series. Each pipeline has two convolutional layers with ReLU activation and max pooling. The outputs are flattened, concatenated, and fed into a fully connected layer with 732 neurons, followed by a softmax output layer. The model is compiled with categorical cross-entropy loss and optimized using SGD. This architecture captures dependencies and interactions in multivariate time series data.

\subsubsection{Time convolutional neural network (Time-CNN)}
The Time Convolutional Neural Network is designed for time series classification, distinguishing itself through its unique architecture and choice of operations. The model has two convolutional layers with 6 and 12 filters, using sigmoid activation, followed by average pooling. Unlike traditional CNNs, it uses mean squared error (MSE) as the loss function. The final classification layer is a fully connected layer with sigmoid activation, not restricting the output to a probability distribution.

\subsubsection{Time warping invariant echo state network (TWIESN)}
The TWIESN model classifies time series data using echo state networks. It initializes sparse, randomly connected weight matrices, selecting parameters like units and spectral radius based on predefined configurations. A state matrix is computed for each time step, combining current input and previous state with non-linear activation and a leaky integrator. These states and the original input form a new feature space. Training uses Ridge regression with hyperparameters selected via grid search. For testing, predictions are made by computing the state matrix and averaging over time steps.

\subsubsection{Inception}
The InceptionTime algorithm, inspired by the Inception architecture, handles time series data variability using multiple convolutional filters of different lengths within the same layer. This multi-scale approach captures diverse features at different resolutions. The model includes bottleneck layers for dimensionality reduction and residual connections to improve training and performance. It features a global average pooling layer and a dense softmax output layer for robust classification, trained with categorical cross-entropy loss and optimized with Adam.

\subsubsection{ConvTran}
The ConvTran model combines convolutional layers and transformer blocks to classify multivariate time series data. It starts with convolutional layers to extract spatial and temporal features, using batch normalization and GELU activation. Positional encoding (e.g., tAPE, Absolute, Learnable, Attention\_Rel\_Scl, and Attention\_Rel\_Vec) retains temporal order. The transformer block employs multi-head attention to capture long-range dependencies, with LayerNorm and a Feed-Forward network for further processing. Final layers include global average pooling, flattening, and a fully connected output layer for classification.

\subsection{DL and Transformer Model Validation}\label{subsec_model_validation}

In the evaluations presented later in this paper, these DL and ConvTran methods do not perform very well. So to confirm that our implementations are working correctly we have conducted a validation exercise to compare against some available benchmarks. Since our AcTBeCalf data is multivariate the validation of the DL models considers three multivariate datasets: 
CharacterTrajectories \citep{williams2006extracting}, JapaneseVowels \citep{kudo1999multidimensional}, Libras \citep{dias2009hand}.
One univariate dataset, Wafer \citep{olszewski2001generalized} is also included. 
Baseline results on the WISDM dataset \citep{weiss2012impact} are included for the validation of the ConvTran model. The objective of these validations is to reproduce the results reported by \citet{ismail2019deep} for the DL models and by \citet{ConvTran2023} for the ConvTran model. 

\begin{figure}[hbtp]
\centering
\includegraphics[width=0.95\textwidth]
{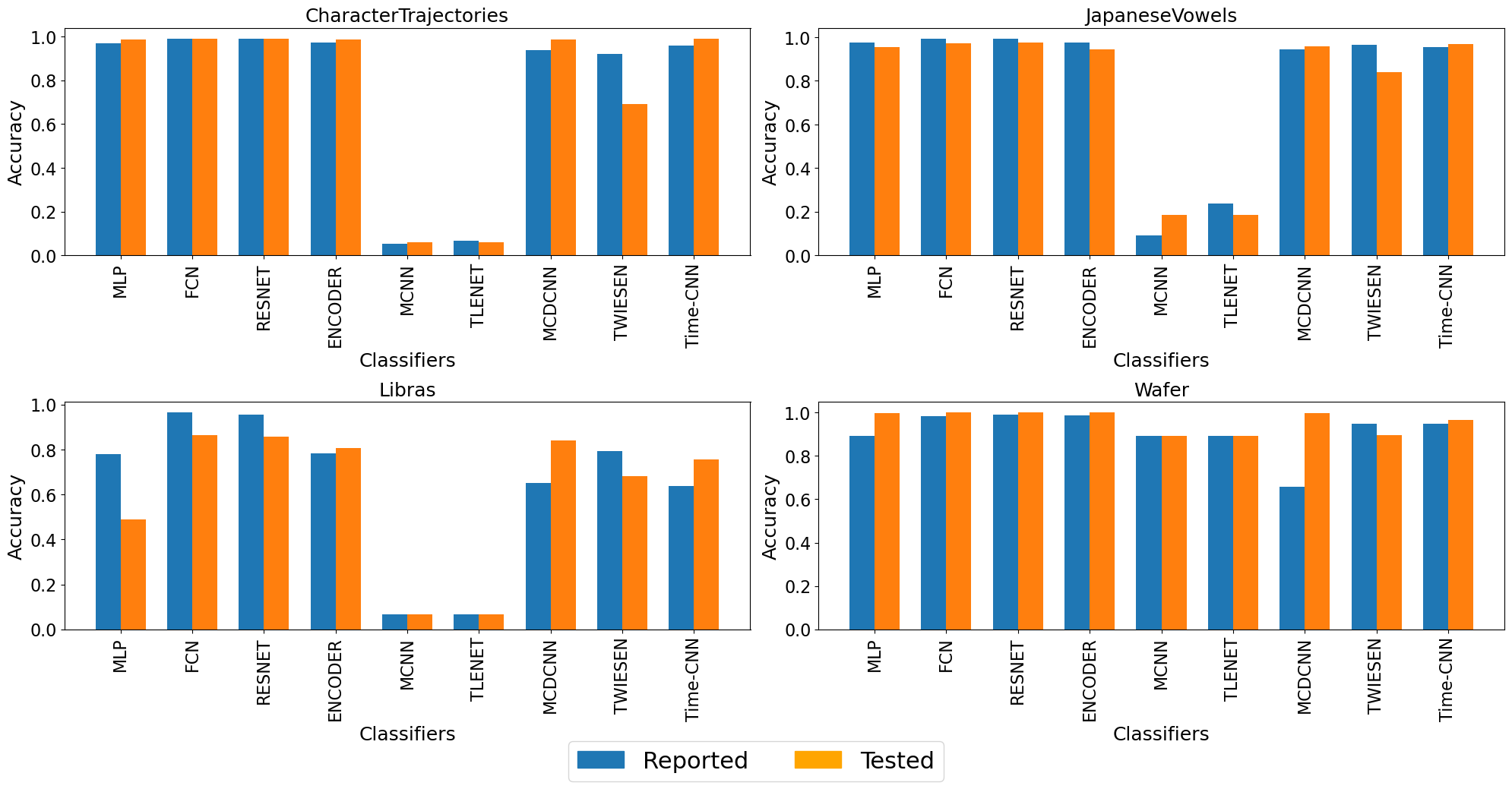}
\caption{A comparison of accuracies of our DL model implementations against some published baselines.}
\label{fig:accuracy_DL_validation}
\end{figure}

Figure \ref{fig:accuracy_DL_validation} presents the accuracy of the DL models, comparing the reported results by \citet{ismail2019deep} with our tested results. The tested results align closely with the results reported. The very poor performance of MCNN and t-LeNet in both the baseline and our validation tests stand out. \citet{ismail2019deep} suggest that this is probably because 
these approaches  extract sub-sequences to augment the training data and these sub-sequences can miss  discriminating information. 

\citet{ConvTran2023} has reported an accuracy of 0.9098 for the ConvTran model with the WISDM dataset and our implimentation of the model also produces an accuracy score of 0.91\%.


The code used for data splitting, performing the time series classification using ROCKET, ConvTran and DL methods is available in the GitHub repository \footnote{https://github.com/Oshana/ESWA}.

\section{Dataset}\label{sec_dataset}

\subsection{Data Collection and Dataset Summary}\label{subsec_data_collection}

\begin{figure}[hbtp]
\centering
\includegraphics[width=0.95\textwidth]
{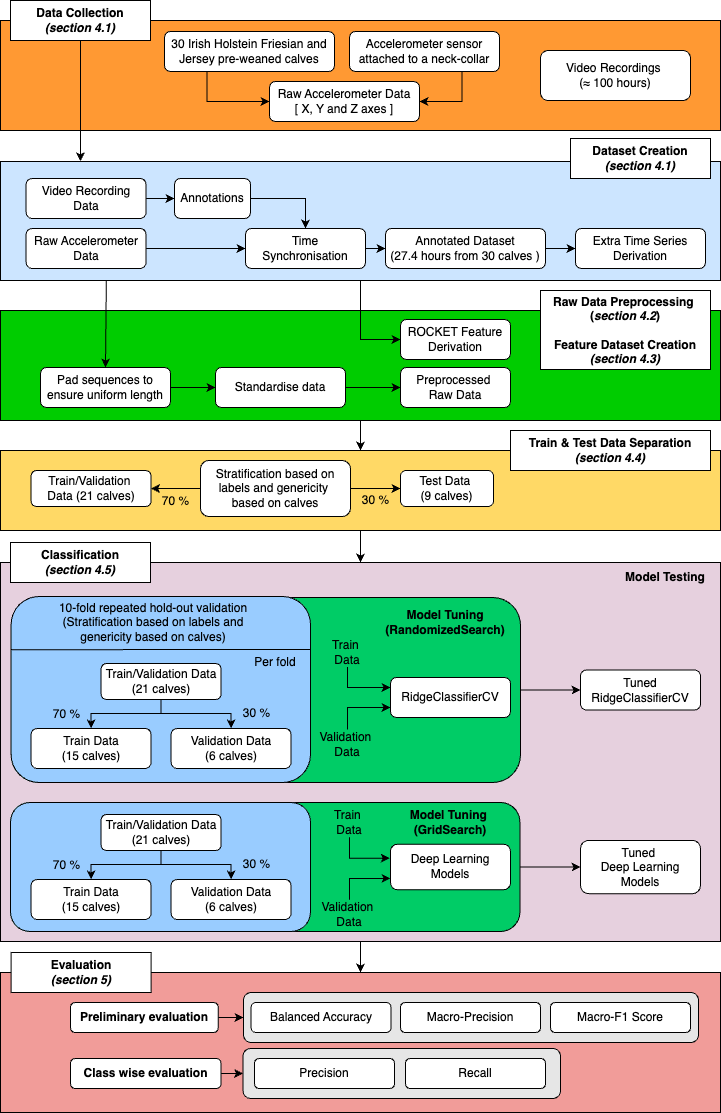}
\caption{Overview of the methodology used to compare classification performance and to assess generalisation capability across calves.}
\label{fig:pipeline}
\end{figure}

A high-level visualisation of the methodology followed is shown in Figure \ref{fig:pipeline}.

A detailed description including the data collection and dataset creation of the ActBeCalf dataset used in this study is provided in the paper \citep{dissanayake4928636accelerometer}. Table \ref{tab:behaviour-definitions} outlines the behaviours considered in this study, while Table \ref{tab:behaviour-summary} summarizes the specifics of these behaviours in the ActBeCalf dataset.

\begin{table}[hbtp]
\caption{Definition of the behaviours based on the ethogram used for that study \citep{barry2019development} }
\label{tab:behaviour-definitions}
\begin{tabular}{@{}ll@{}}
\toprule
\multicolumn{1}{c}{Behaviour} & \multicolumn{1}{c}{Definition}                                         \\ \midrule
Drinking milk              & Calf is drinking milk from the milk feeder.                          \\ \midrule
Grooming                     & Calf uses tongue to repeatedly lick own back, side, leg, tail areas. \\ \midrule
Lying & \begin{tabular}[c]{@{}l@{}}Calf is resting either sternally or laterally with all four legs \\ hunched close to body either awake or asleep.\end{tabular} \\ \midrule
Running                      & Calf is running (play / not-play).                                                     \\ \midrule
Walking                      & Calf is walking or shuffling about.                                  \\ \midrule
\textit{Other} & \begin{tabular}[c]{@{}l@{}}A collection of various other behaviours, including rising, lying down, \\social interaction, play, and more.\end{tabular}                   \\ \bottomrule
\end{tabular}
\end{table}

\begin{table}[hbtp]
\caption{Summary of data collected after remapping into 6 classes}
\label{tab:behaviour-summary}
\begin{tabular}{@{}rrrr@{}}
\toprule
Behaviour       & Total Duration (minutes) & Number of segments & Number of calves \\ \midrule
Drinking\ milk & 143.68                  & 169                 & 27               \\
Grooming        & 75.26                   & 334                 & 29               \\
Lying           & 650.36                  & 120                 & 27               \\
Running         & 55.10                   & 608                 & 24               \\
Walking         & 44.29                   & 561                 & 30               \\
\textit{Other}           & 671.63                  & 2636                & 30               \\ \bottomrule
\end{tabular}
\textit{Number of segments: The total count of continuous observations per behaviour recorded by an annotator.}
\end{table}

\subsection{Raw Data Preprocessing}\label{subsec_data_preproc}
The preprocessing of raw data for DL techniques involves two key steps. First, the time series are converted into a uniform size. Given the accelerometer's frequency of 25 Hz and a time window of 3 seconds, each sequence is standardized to 75 data points. This uniformity organizes the data into a consistent shape, suitable for batch processing in deep learning models, ensuring that each sequence in a batch has the same length and thus facilitating efficient and accurate learning from the multivariate time series data. Second, the time series data is standardized. This step involves removing the mean and scaling to unit variance, ensuring that each time series contributes equally to the model's learning process and enhancing the convergence speed of the training algorithm \citep{lecun2002efficient, sutskever2013importance}.

\subsection{ROCKET Feature Derivation}\label{subsec_feature_dataset_creation}
In place of the randomness of ROCKET as explained in Section \ref{subsec_rocket}, we used MiniRocket for this study; this makes use of a more streamlined set of convolutional kernels and focuses on a predetermined subset of kernel parameters. MiniRocket also focuses on identifying and selecting the best features. As a result, MiniRocket can achieve comparable or even better classification performance than ROCKET with a much smaller computational effort \citep{dempster2021minirocket}. 

MiniRocket creates a single feature vector by applying random convolutional kernels to each of the 8 time series individually, generating features for each series. This initially creates 79,968 features (9,996 features per time series; 9,996 features = the nearest multiple of 84 less than 10,000 (the default number of number of kernels) \citep{dempster2021minirocket}). Each convolution operation extracts features from the individual series, which are then pooled using techniques like global average pooling. These pooled features are concatenated to form a comprehensive feature vector that captures the combined information and interactions from all the series, resulting in a unified representation of the entire multivariate time series dataset. 

\subsection{Assessing Generalisation Performance}\label{subsec_data_separation}

\begin{figure}[H]
\centering
\includegraphics[width=0.9\textwidth]
{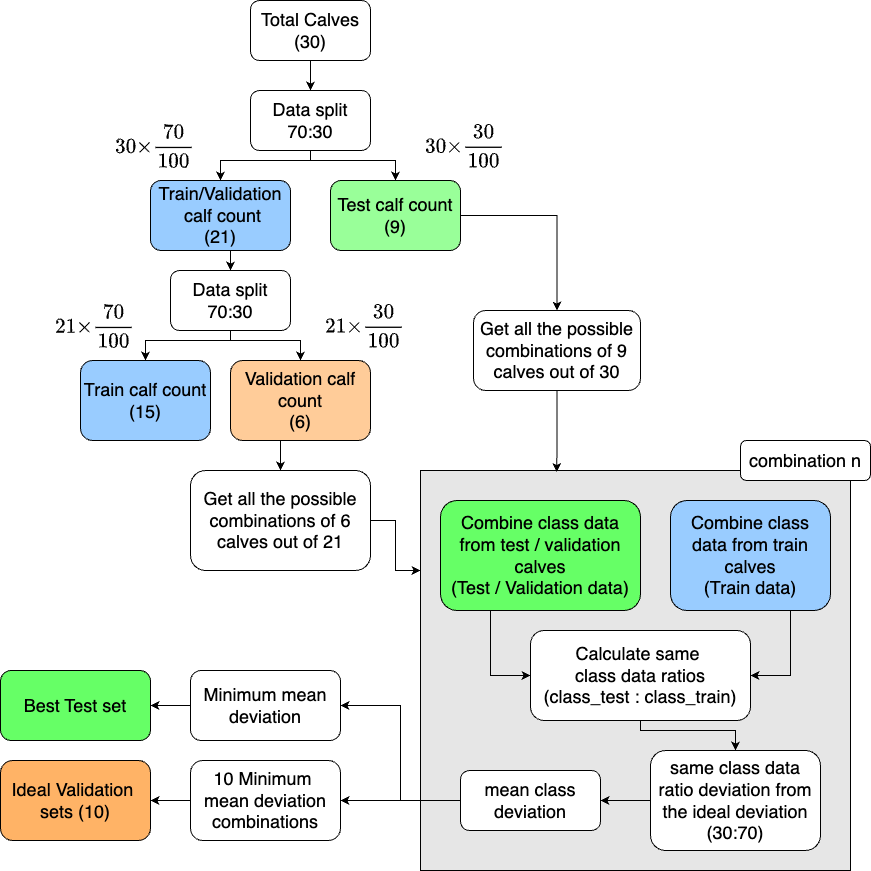}
\caption{Methodology followed to ensure data generalisation.}
\label{fig:generalisation-pipeline}
\end{figure}

Given that our ultimate objective would be to develop models that will generalise to new calves not encountered by the model, for model development and evaluation purposes the data is separated at the calf level (first, the calves are divided into training, testing, and validation groups, and then the data from each calf within these groups is compiled. This approach ensures that each dataset contains data from distinct, non-overlapping calves). In contrast, some studies in the literature mix the data initially and use a simple split ratio to divide into training and testing sets \citep{carslake2020machine, white2008evaluation, zhang2024analysis}. This approach will often result in data from the same subject appearing in both sets, compromising the assessment of generalization performance. Mixing the data at the outset simplifies class stratification and facilitates functions like grid search, as it avoids the need to consider subject-level splits for each cross-validation fold - but it results in over-optimistic assessments of generalisation accuracy. 
Figure \ref{fig:generalisation-pipeline} shows the overall methodology used to achieve the above goals.
\\
The study employs a 30:70 split ratio for test and training data to ensure sufficient data for model testing. This approach is applied to both the test and validation splits. Initially, all possible combinations of 9 out of 30 calves ($9 = 0.3\times 30$) are generated. For each combination, the test dataset is formed using data from the calves in that combination, while the remaining 21 calves ($21 = 0.7 \times 30$) are used for training.

Next, the class data ratios are calculated, and their deviation from the ideal ratio of 0.43 (30/70) is determined for each class. This step helps assess how well the combinations maintain class-level data distribution (stratification). The mean value of these deviations is computed for each combination; a lower mean value indicates better stratification of class data. This process is repeated for all combinations, and the combination with the smallest deviation is selected as the ideal test set to ensure class-level stratification.

Using the remaining 21 calves, the same procedure is followed for all possible combinations of 6 calves ($6 = 0.3 \times 21$: validation sets). The combinations with the minimum deviation are chosen for the number of folds required, which is 10 in this study.

\subsection{Classification}\label{subsec_classification}

A grid search is employed to determine the optimal hyperparameters for the Ridge Classifier using ROCKET features. The  hyperparameters considered are listed in Table \ref{tab:RCV-hyperparams}. To ensure the generalization of the validation data, the data is partitioned at the calf level for each fold, as detailed in Section \ref{subsec_data_separation}. Indexes for the training set (comprising 15 calves) and the validation set (comprising 6 calves) per fold are predetermined and stored for the grid search. This index corpus is provided as custom cross-validation information for the grid search. 50 different hyper-parameter combinations were evaluated using 10 fold cross-validation. The code for this is available in the GitHub repository. The best hyperparameters identified are:

\begin{itemize}
    \item \textbf{alpha: } 131.31
    \item \textbf{class\_weight: } balanced
    \item \textbf{fit\_intercept: } False
\end{itemize}

\begin{table}[hbtp]
\caption{RidgeClassifierCV - Tested Hyperparameters}
\label{tab:RCV-hyperparams}
\begin{tabular}{@{}ll@{}}
\toprule
Hyperparameter          & Values                                            \\ \midrule
\textbf{alphas}         & 100 linearly spaced values between 0.001 and 1000 \\
\textbf{class\_weight}  & None, balanced                                    \\
\textbf{fit\_intercept} & True, False                                       \\ \bottomrule
\end{tabular}
\end{table}

For the deep learning techniques, the data was partitioned into test, validation, and training sets. The test set consisted of the 9 calves with the best label stratification, followed by the 6 calves with the best label stratification from the remaining 21 calves. The remaining 15 calves were used as the training set. Certain of the DL techniques, such as MCNN, MCDCNN, and Twi-ESN, were adapted to use the data separated at the calf level instead of employing the internal train\_test\_split function from scikit-learn, ensuring consistent generalisation across all DL techniques. The same set of test, validation, and training data was also used for ConvTran.

\section{Evaluation}\label{sec_evaluation}


\begin{figure}[hbtp]
\centering
\includegraphics[width=1\textwidth]
{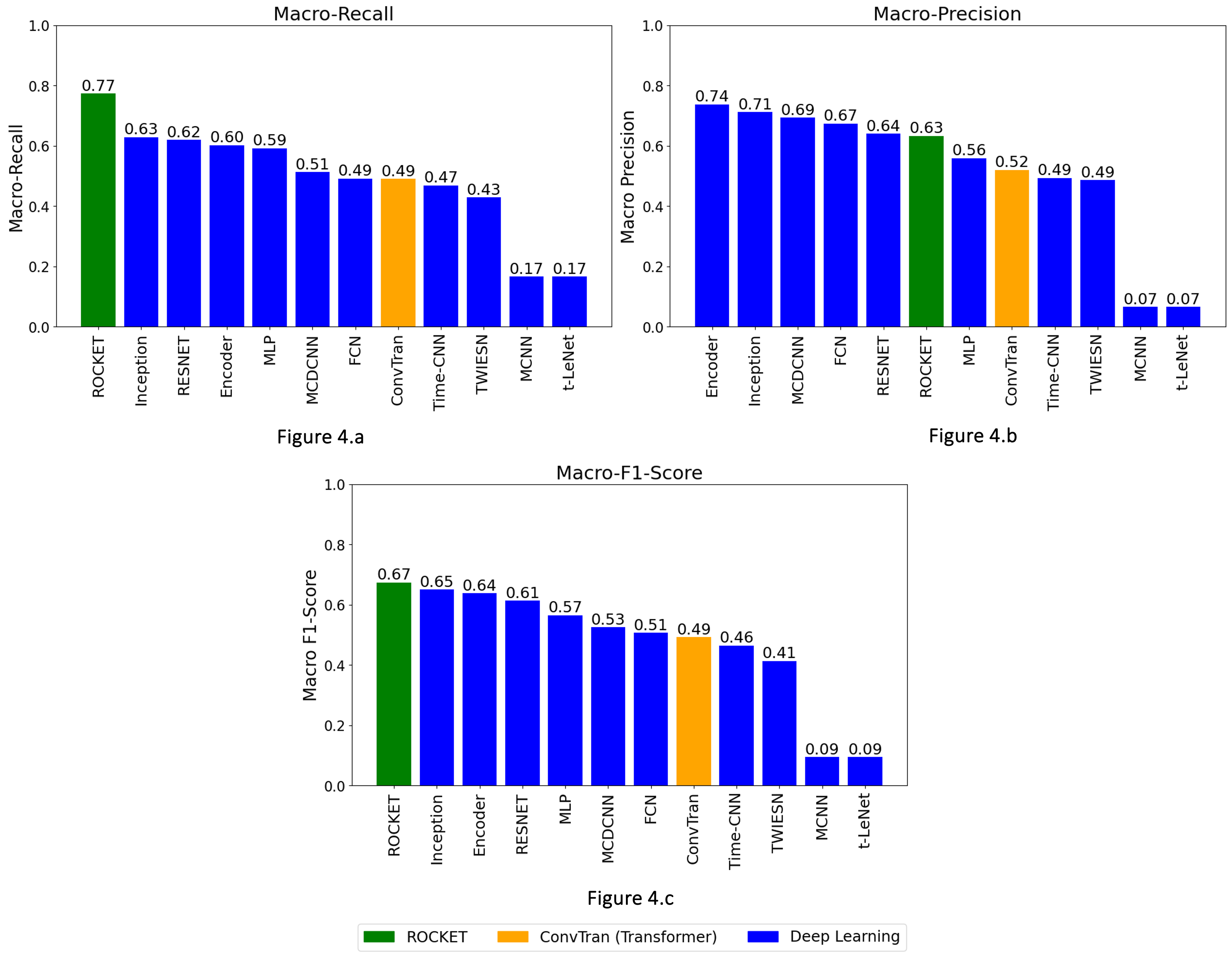}
\caption{Macro-recall, macro-precision, and macro-F1 scores for each of the models}
\label{fig:all_model_results}
\end{figure}

Given that the overall objective is to get an accurate inventory of calf behaviours, we use Precision and Recall as the key measures in the evaluation. Precision quantifies that proportion of labels assigned for a class that are correct and Recall quantifies the proportion of actual events that are picked up. Given that the data is multi-class macro-precision and macro-recall figures are reported. These are averages of the scores for each of the labels. 

 Figure \ref{fig:all_model_results}.a presents the macro-recall scores for all models. Among the models, the ROCKET algorithm exhibited the highest macro-recall at 0.77, significantly outperforming the DL and ConvTran (transformer) methods. The Inception and ResNet DL models demonstrated comparatively better performance with scores of 0.63 and 0.62, respectively. Conversely, Encoder, MLP, and MCDCNN models achieved moderate Recall, ranging from 0.60 to 0.51. The transformer-based model, ConvTran, and other DL methods, like FCN, Time-CNN, and TWIESN, showed lower Recall scores between 0.49 and 0.43. Notably, the MCNN and t-LeNet models had the lowest Recall at 0.17. 


The macro-precision scores are presented by the Figure \ref{fig:all_model_results}.b. The Encoder model achieved the highest macro-precision score at 0.74. The Inception and MCDCNN models followed with scores of 0.71 and 0.69, respectively. Models such as FCN, ResNet, and ROCKET showed moderately high Precision, ranging from 0.67 to 0.63. The MLP and ConvTran models scored 0.56 and 0.52, while Time-CNN and TWIESN scored 0.49. The MCNN and t-LeNet models had the lowest macro-precision at 0.07.




Finally, the F1 scores are shown in Figure \ref{fig:all_model_results}.c. Again ROCKET achieved the highest macro F1-score at 0.67, indicating its better balance between precision and recall across all classes. The Inception and Encoder models followed closely with scores of 0.65 and 0.64, respectively. Other models, including ResNet, MLP, and MCDCNN, exhibited moderate F1-scores ranging from 0.61 to 0.53. The ConvTran model and other DL methods like FCN, Time-CNN, and TWIESN displayed lower F1-scores, with values between 0.51 and 0.41. Notably, the MCNN and t-LeNet models had the lowest macro F1-scores at 0.09. 


\subsection{Performance on Class Labels}\label{subsec_class_wise_eval}

Given that this is a multi-class problem, we selected the most promising models for more detailed analysis. 
We consider ROCKET and the two best performing DL models, Inception and Encoder. Despite its lower performance we also consider ConvTran because of its interesting architecture, which integrates CNN and transformers, offering insightful comparison to both ROCKET and other DL models.

\begin{figure}[H]
\centering
\includegraphics[width=0.82\textwidth]
{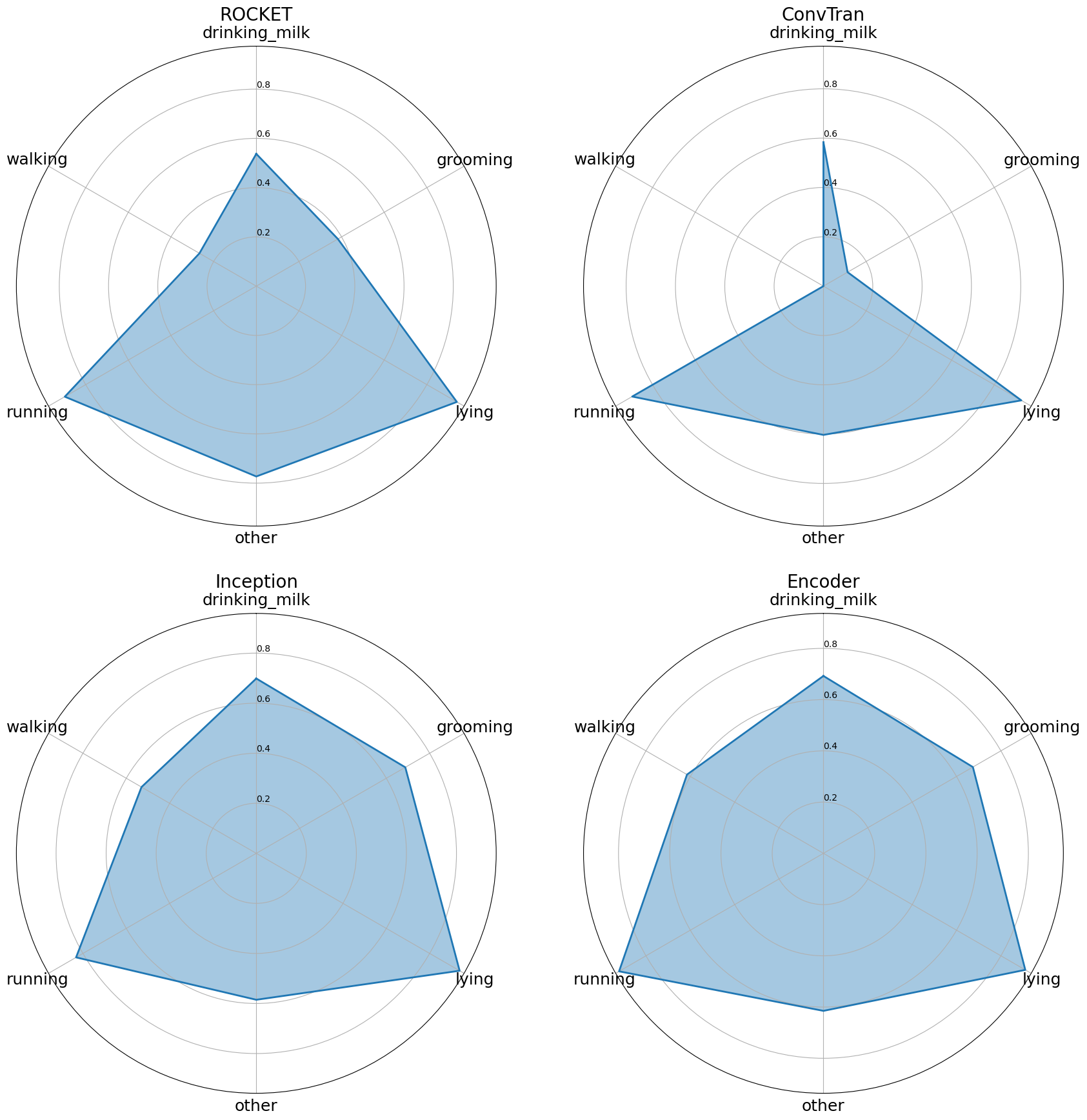}
\caption{Class-level Precision for ROCKET, ConvTran the top performing DL algorithms.}
\label{fig:class_wise_precision}
\end{figure}

Figure \ref{fig:class_wise_precision} illustrates the model precision across the class labels. 
ROCKET demonstrates high precision in identifying running, lying, and \textit{other} behaviours and moderate performance for the grooming, walking and drinking milk behaviours. In contrast, ConvTran performs poorly in the walking and grooming classes but performs well in the running and lying classes. Among the DL models, Inception shows really good performance in all classes except moderate performance for the \textit{other} class. The Encoder model has the best overall-performance and perform best for the walking and running classes but moderate performance for the \textit{other} class. Overall, Encoder exhibits the best performance, followed by Inception and ROCKET, while ConvTran performs the poorest.

\begin{figure}[hbtp]
\centering
\includegraphics[width=0.82\textwidth]
{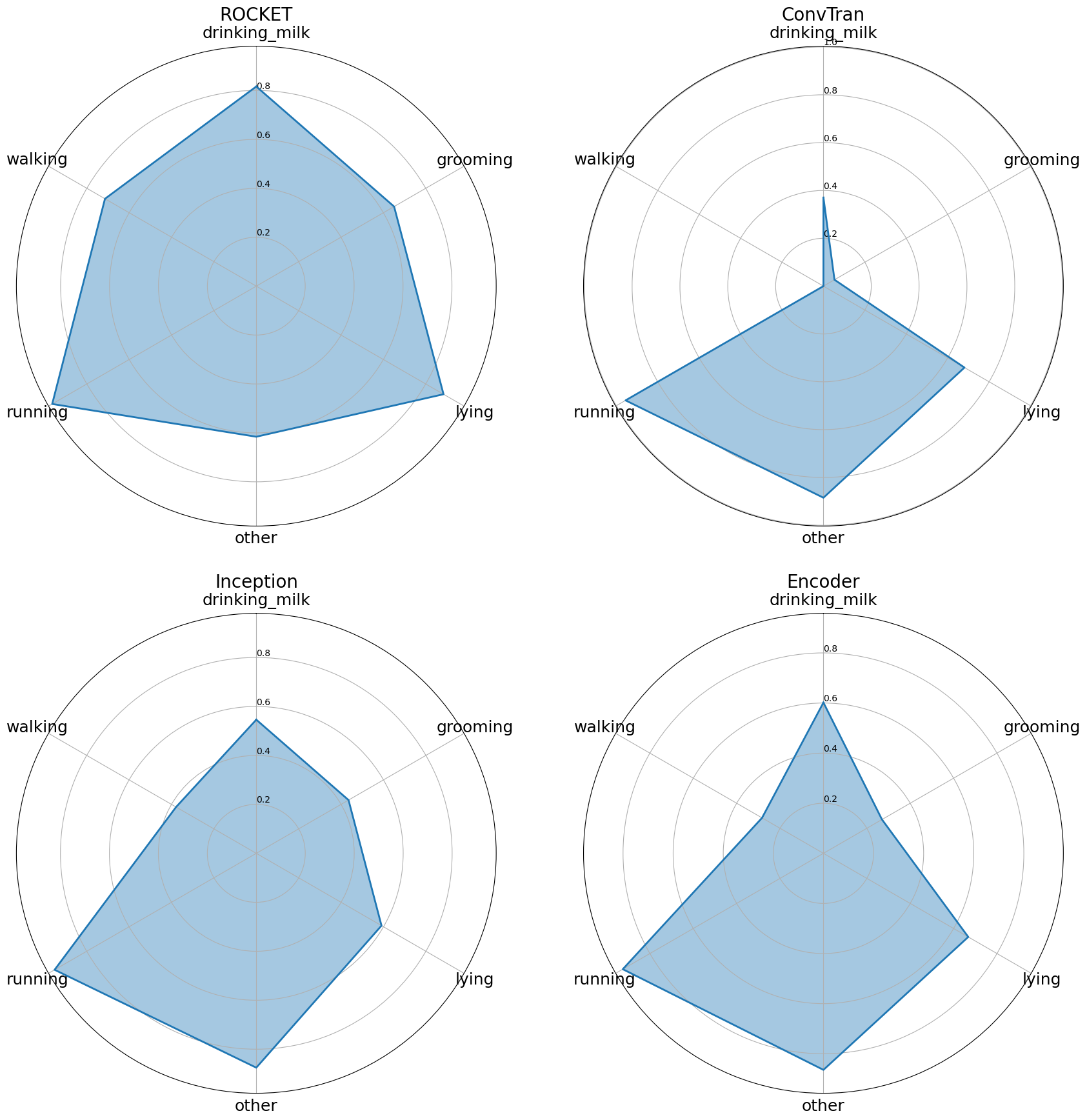}
\caption{Class-level Recall for ROCKET, ConvTran the top performing DL algorithms.}
\label{fig:class_wise_recall}
\end{figure}

Figure \ref{fig:class_wise_recall} presents the recall figures for individual labels for ROCKET, ConvTran, and the best DL models in the study. Overall ROCKET shows the best performance with high recall values for walking, drinking milk and grooming which is important in terms of less prominent behaviour identification. ROCKET also shows exceptional performance for the running and lying classes. ConvTran model continues to perform poorly except for the running and \textit{other} classes. Inception and Encoder perform well for the running and \textit{other} classes and performs moderately for the lying, walking and drinking milk classes. Encoder has lower performance for the grooming class than the Inception.

\begin{figure}[hbtp]
\centering
\includegraphics[width=\textwidth]
{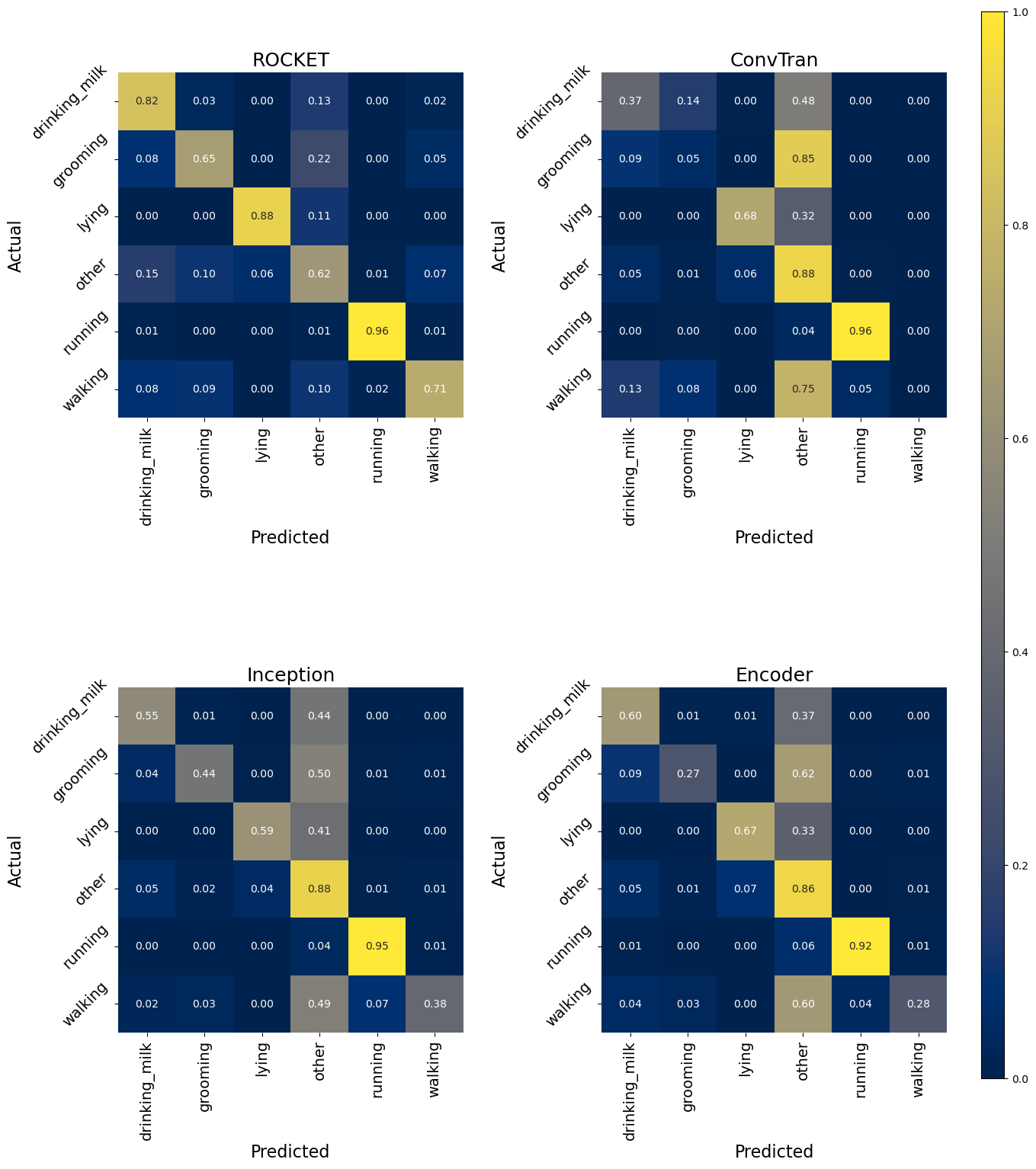}
\caption{Confusion Matrices for ROCKET, ConvTran the top performing DL algorithms.}
\label{fig:cms}
\end{figure}

The confusion matrices in Figure \ref{fig:cms} provide a detailed comparison of the performance of the selected algorithms. ROCKET's confusion matrix shows strong performance with high true positive rates across most behaviour classes. 
In contrast the confusion matrices for ConvTran, Inception and Encoder show that they have problems with the \emph{other} class. This problem is most evident with ConvTran. Inception shows balanced performance, with moderate mis-classifications spread evenly across different classes but still performs well for the running and \textit{other} class. The Encoder model accurately identifies specific classes like running and \textit{other} but struggles more with the rest of the activities. 

\section{Discussion}\label{sec_discussion}

The main finding from this study is that  ROCKET significantly outperforms DL and transformer-based (ConvTran) methods in classifying calf behaviours. ROCKET achieved the highest macro-recall of 0.77, markedly higher than the best-performing DL methods, such as Inception, ResNet and Encoder, which achieved balanced accuracies of 0.63, 0.62 and 0.6, respectively. Given the enthusiasm for DL methods these days this is an interesting result; however it is not completely unexpected because a recent assessment of DL algorithms in TSC came up with similar findings \citep{middlehurst2024bake}.

While ROCKET performed best in terms of macro-recall and the macro F1 score it is worth emphasising that some of the DL methods (particularly the Encoder and Inception model) had better macro-precision.
This is important for minimising false positives in behavioural classification.



Several factors may contribute to the superior performance of ROCKET compared to DL methods. ROCKET’s feature extraction process leverages a large number of random convolutional kernels, which can capture a wide variety of patterns in the data without the need for extensive training. This results in a high-dimensional feature space well-suited for linear classifiers, such as the Ridge Classifier, which is efficient and effective for this data type. In contrast, DL methods, including convolutional neural networks (CNNs) and recurrent neural networks (RNNs), often require large amounts of data to generalise well. Despite being substantial, the dataset used in this study might still be insufficient for these complex models to fully capture the complexities of calf behaviours.
Furthermore, overfitting of DL models is common, particularly in the case of imbalanced training data. This can lead to reduced performance on the test set. The transformer-based method, ConvTran, also underperformed compared to ROCKET, with a macro-recall score of 0.49. ConvTran, which integrates convolutional layers with transformer blocks, relies heavily on its ability to model long-range dependencies within the data. However, its performance might be hindered by the relatively short and fragmented nature of the time-series data in this study. It will be insightful to evaluate how the performance of behaviour classification is affected by the use of longer duration segments.

The results underscore the potential of methods such as ROCKET that incorporate feature extraction in animal behaviour classification tasks. The high performance of ROCKET in this study shows that it can be a valuable and practical tool in precision livestock farming, enabling accurate and efficient monitoring of calf behaviours. 
Despite the promising results, more data is required to mitigate the imbalance  between the classes. Data augmentation mechanisms also could mitigate this imbalance nature. Other important but less frequent behaviours require further investigation and may benefit from different modelling approaches. 
For instance, using ROCKET for initial feature extraction followed by DL for classification might yield better results. 

In conclusion, the comprehensive analysis of various performance metrics establishes ROCKET as this study’s most effective algorithm for behaviour classification. While DL models like Inception, Encoder and, ResNet show promise, they cannot match ROCKET’s consistent and robust performance. These findings suggest that future research should focus on optimising DL and transformer-based models to potentially close the performance gap and further improve classification accuracy and reliability.

\section{Conclusion}\label{sec_conclusion}

This study evaluated DL, ConvTran and ROCKET for classifying calf behaviours using accelerometer data. ROCKET  emerged as the top performer, achieving a macro-recall of 0.77, significantly higher than the best DL models, such as Inception, and Encoder, which achieved balanced accuracies of 0.63 and, 0.6 respectively. ROCKET's superior performance is attributed to its robust feature extraction process, which captures a wide range of temporal patterns without requiring extensive data or complex training.

While demonstrating high precision and recall in specific contexts, DL models generally struggled with the complexity and variability of the calf behaviour dataset. The transformer-based ConvTran model also underperformed, likely due to the fragmented nature of the time-series data used in this study.

Class-level analysis revealed that ROCKET consistently maintained high true positive rates across most behaviour classes, with exceptional performance in identifying behaviours such as running, lying, drinking milk and grooming. Conversely, DL models exhibited more variability in their performance. The recall analysis showed that all models performed reasonably well for the classes running and lying. Only ROCKET was able to achieve good recall values for walking, drinking milk and grooming.

These findings underscore the potential of established machine learning methods like ROCKET in achieving high accuracy and reliability in animal behaviour classification tasks. While deep learning models have shown promise, there is a need for further optimization and exploration to bridge the performance gap.

Future research should focus on enhancing data representation and increasing dataset size to improve the generalisability of deep learning models. Additionally, exploring hybrid models that combine the strengths of both traditional and DL approaches could offer new avenues for improving classification performance. Developing more accurate and efficient behaviour monitoring systems will significantly enhance animal welfare and farm productivity by enabling timely interventions and better management practices.

\section*{Author contributions: CRediT}
\textbf{Oshana Dissanayake: }Conceptualization, Data curation, Methodology, Software, Writing – original draft. \textbf{Sarah E. McPherson: } Data collection and curation. \textbf{Emer Kennedy: } Data collection and curation. \textbf{Lucile Riaboff: } Data collection and curation, Writing – review and editing. \textbf{Pádraig Cunningham: }Conceptualization, Methodology, Supervision, Writing – review and editing.

\section*{Acknowledgements} 
This publication has emanated from research conducted with the financial support of SFI and the Department of Agriculture, Food and Marine on behalf of the Government of Ireland to the VistaMilk SFI Research Centre under Grant Number 16/RC/3835.

\section*{Declaration of competing interests}
The authors declare that they have no known competing financial interests or personal relationships that could have appeared to influence the work reported in this paper.

\newpage
\begin{appendices}

\section{Precision and Recall for each model}\label{secA1}

\begin{table}[ht]
\label{tab:legend}
\begin{tabular}{@{}
>{\columncolor[HTML]{FBBC04}}l r@{}}
 & Highest Score
\end{tabular}
\end{table}

\begin{table}[ht]
\caption{Precision values scored by the models}
\label{tab:app-all-precision}
\begin{tabular}{@{}lcccccc@{}}
\toprule
\multicolumn{1}{c}{Model} &
  \textbf{Drinking Milk} &
  \textbf{Grooming} &
  \textbf{Lying} &
  \textbf{Running} &
  \textbf{Walking} &
  \textbf{Other} \\ \midrule
\textbf{ROCKET}    & 0.54 & 0.38                         & \cellcolor[HTML]{FBBC04}0.94 & 0.90 & 0.27 & \cellcolor[HTML]{FBBC04}0.77 \\
\textbf{ConvTran}  & 0.58 & 0.11                         & 0.93                         & 0.90 & 0.00 & 0.60                         \\
\textbf{Inception} & 0.70 & \cellcolor[HTML]{FBBC04}0.69 & \cellcolor[HTML]{FBBC04}0.94 & 0.83 & 0.53 & 0.59                         \\
\textbf{Encoder}   & 0.69 & 0.67                         & 0.91                         & 0.92 & 0.61 & 0.61                         \\
FCN &
  \cellcolor[HTML]{FBBC04}0.71 &
  0.22 &
  0.92 &
  \cellcolor[HTML]{FBBC04}0.99 &
  \cellcolor[HTML]{FBBC04}0.69 &
  0.52 \\
Time-CNN           & 0.45 & 0.04                         & 0.85                         & 0.89 & 0.14 & 0.59                         \\
TWIESN             & 0.59 & 0.00                         & 0.77                         & 0.94 & 0.00 & 0.61                         \\
RESNET             & 0.59 & 0.20                         & 0.88                         & 0.94 & 0.60 & 0.63                         \\
MCNN               & 0.00 & 0.00                         & 0.00                         & 0.00 & 0.00 & 0.40                         \\
t-LeNet            & 0.00 & 0.00                         & 0.00                         & 0.00 & 0.00 & 0.40                         \\
MCDCNN             & 0.69 & 0.39                         & 0.92                         & 0.95 & 0.60 & 0.61                         \\
MLP                & 0.55 & 0.21                         & 0.90                         & 0.90 & 0.19 & 0.60                         \\ \bottomrule
\end{tabular}
\end{table}

\begin{table}[ht]
\caption{Recall values scored by the models}
\label{tab:app-all-recalls}
\begin{tabular}{@{}lcccccc@{}}
\toprule
\multicolumn{1}{c}{Model} &
  \textbf{Drinking Milk} &
  \textbf{Grooming} &
  \textbf{Lying} &
  \textbf{Running} &
  \textbf{Walking} &
  \textbf{Other} \\ \midrule
\textbf{ROCKET} &
  \cellcolor[HTML]{FBBC04}0.82 &
  \cellcolor[HTML]{FBBC04}0.65 &
  \cellcolor[HTML]{FBBC04}0.88 &
  \cellcolor[HTML]{FBBC04}0.96 &
  \cellcolor[HTML]{FBBC04}0.71 &
  0.62 \\
\textbf{ConvTran}  & 0.37 & 0.05 & 0.68                         & \cellcolor[HTML]{FBBC04}0.96 & 0.00 & 0.88                         \\
\textbf{Inception} & 0.55 & 0.44 & 0.59                         & 0.95                         & 0.38 & 0.88                         \\
\textbf{Encoder}   & 0.60 & 0.27 & 0.67                         & 0.92                         & 0.28 & 0.86                         \\
FCN                & 0.39 & 0.36 & 0.35                         & 0.82                         & 0.12 & 0.91                         \\
Time-CNN           & 0.46 & 0.11 & 0.51                         & 0.94                         & 0.03 & 0.77                         \\
TWIESN             & 0.03 & 0.00 & \cellcolor[HTML]{FBBC04}0.88 & 0.92                         & 0.00 & 0.75                         \\
RESNET             & 0.57 & 0.42 & 0.60                         & 0.94                         & 0.39 & 0.79                         \\
MCNN               & 0.00 & 0.00 & 0.00                         & 0.00                         & 0.00 & \cellcolor[HTML]{FBBC04}1.00 \\
t-LeNet            & 0.00 & 0.00 & 0.00                         & 0.00                         & 0.00 & \cellcolor[HTML]{FBBC04}1.00 \\
MCDCNN             & 0.47 & 0.05 & 0.71                         & 0.94                         & 0.02 & 0.89                         \\
MLP                & 0.52 & 0.34 & 0.69                         & 0.95                         & 0.36 & 0.70                         \\ \bottomrule
\end{tabular}
\end{table}

\end{appendices}

\clearpage


\bibliography{sn-bibliography}


\end{document}